\documentclass{article}

\usepackage[utf8]{inputenc}
\usepackage{scicite}
\usepackage[margin=1in]{geometry}
\usepackage[T1]{fontenc}
\usepackage{url}
\usepackage{booktabs}
\usepackage{amsfonts}
\usepackage{nicematrix}
\usepackage{graphicx}
\usepackage{pgfplots}
\usepackage{cite}
\usepackage{hyperref}
\usepackage{float}

\title{ARC-AGI-2:
A New Challenge for Frontier AI Reasoning Systems}

\author{
    Fran\c{c}ois Chollet\thanks{fchollet@arcprize.org} \and
    Mike Knoop \and
    Gregory Kamradt \and
    Bryan Landers \and
    Henry Pinkard \and
}

\begin{document}

\maketitle

\begin{abstract}
The Abstraction and Reasoning Corpus for Artificial General Intelligence (ARC-AGI), introduced in 2019, established a challenging benchmark for evaluating the general fluid intelligence of artificial systems via a set of unique, novel tasks only requiring minimal prior knowledge. While ARC-AGI has spurred significant research activity over the past five years, recent AI progress calls for benchmarks capable of finer-grained evaluation at higher levels of cognitive complexity. We introduce ARC-AGI-2, an upgraded version of the benchmark. ARC-AGI-2 preserves the input-output pair task format of its predecessor, ensuring continuity for researchers. It incorporates a newly curated and expanded set of tasks specifically designed to provide a more granular signal to assess abstract reasoning and problem-solving abilities at higher levels of fluid intelligence. To contextualize the difficulty and characteristics of ARC-AGI-2, we present extensive results from human testing, providing a robust baseline that highlights the benchmark's accessibility to human intelligence, yet difficulty for current AI systems. ARC-AGI-2 aims to serve as a next-generation tool for rigorously measuring progress towards more general and human-like AI capabilities.
\end{abstract}

\section{ARC-AGI-1: 2019-2024 history}

The Abstraction and Reasoning Corpus (ARC), later referred to as ARC-AGI-1, was introduced by Fran\c{c}ois Chollet in the 2019 paper ``On the Measure of Intelligence'' \cite{chollet2019measure}. It represented a significant departure from traditional AI benchmarks, which often focused on specific skills or knowledge recall within large datasets. Instead, ARC-AGI-1 was designed to evaluate a more general, human-like form of fluid intelligence --- the ability to reason and solve novel problems efficiently, independent of extensive prior experience or domain-specific training.

\subsection{The dataset}

\begin{figure}[h]
  \centering
  \includegraphics[width=0.8\textwidth]{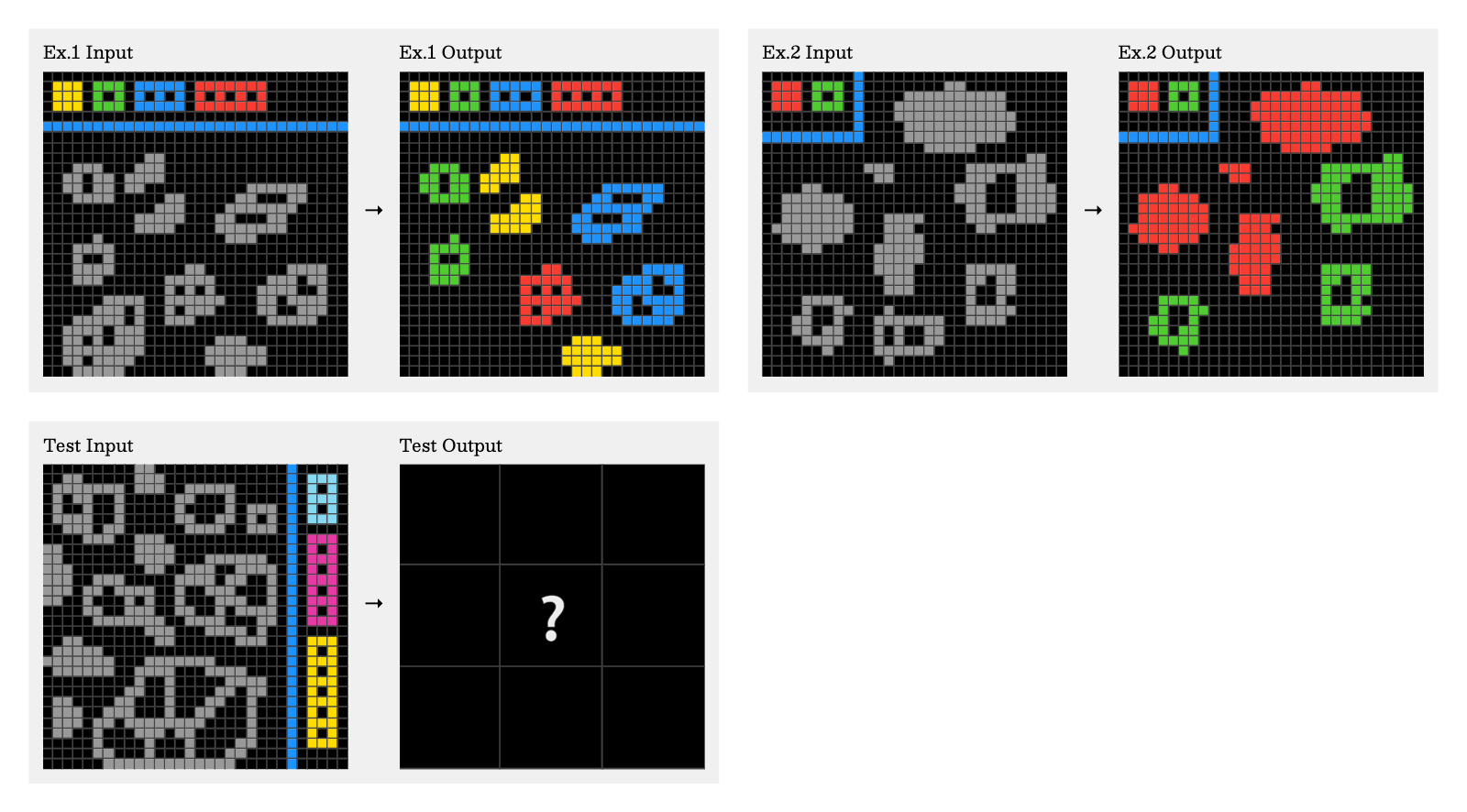}
  \caption{ARC-AGI-2 task id: e3721c99}
  \label{fig:task-e3721c99}
\end{figure}

The dataset consists of a collection of reasoning tasks presented as pairs of grids of discrete symbols (displayed as colored cells). Each task comprises a small number (typically 2-5) of demonstration pairs, where each pair shows an input grid transformed into an output grid according to some underlying, unstated rule. The objective for the test-taker (human or AI) is to infer the rule from these few examples and apply it correctly to one or more unseen test input grids to produce the corresponding output grids. Grid sizes vary but are capped at 30×30, using up to 10 distinct colors. An example can be found in Figure \ref{fig:task-e3721c99}.

The original public dataset contained 400 training tasks and 400 evaluation tasks. A Private Evaluation set of 100 tasks was reserved for competition leaderboards, and a further 100-task "Semi-Private" Evaluation set was introduced in mid-2024 for verifying closed-source models \cite{arcprize:leaderboard}.

ARC-AGI has three defining characteristics:

\begin{enumerate}

\item It aims to resist overfitting and memorization, focusing instead purely on general fluid intelligence. It should not be possible to prepare for any of the tasks in advance. Every task in the dataset is unique.

\item It minimizes the need for prior knowledge. ARC-AGI tasks only require a core set of innate human cognitive priors, such as object persistence, goal-directedness, elementary number sense (counting), and basic geometric and topological concepts (connectivity, symmetry) \cite{chollet2019measure}. Crucially, it requires absolutely no specialized world knowledge (e.g., historical facts) nor language to solve, making it distinct from knowledge-intensive benchmarks commonly used for evaluating LLMs.

\item It is feasible for regular humans without special training.
\end{enumerate}

\subsection{Prior competitions}

The ARC-AGI-1 benchmark has been the focus of several high-profile public competitions, with prize pools increasing significantly over time as interest in the benchmark steadily grew. The competition exists to inspire researchers to work on new ideas that will drive open-source progress on highly efficient, general systems capable of beating ARC-AGI.

Past ARC-AGI competitions (currency in USD):

\begin{enumerate}
    \item 2020: First ARC-AGI Kaggle competition (\$20,000 in prizes)~\cite{kaggle2020}
    \item 2022: ARCathon 1 (\$100,000 in prizes)~\cite{arcathon2022}
    \item 2023: ARCathon 2 (\$100,000 in prizes)~\cite{arcathon2023}
    \item 2024: ARC Prize 2024 (\$1,100,000 in prizes)~\cite{arcprize2024}
\end{enumerate}

For several years following its introduction, progress remained slow. The winning entry in the 2020 competition achieved a score of 20\% on the Private Evaluation set using program synthesis techniques~\cite{kaggle2020}. Despite the rapid scaling of large language models (LLMs) between 2020 and early 2024 (during which base LLMs were scaled up by a factor of more than 10,000x), the state-of-the-art ARC-AGI scores hovered around 34\%~\cite{arcprize2024}, well below estimated human performance.

The landscape shifted rapidly during the ARC Prize 2024 competition, the first competition run by ARC Prize Foundation \cite{arcprizefoundation}. The highest score achieved during the competition on the Private Evaluation set reached 55.5\% (by the MindsAI team), although this solution was not open-sourced and was thus ineligible for prizes. The winning team with an eligible submission, ``the ARChitects'', achieved 53.5\%~\cite{arcprize2024}. This significant jump from the previous state-of-the-art highlighted the effectiveness of novel approaches spurred by the competition, in particular Test-Time Adaptation (TTA) approaches, in which a model dynamically adapts at test time via either test-time search (e.g., chain-of-thought synthesis or symbolic program synthesis) or test-time gradient descent. These approaches are discussed in the ARC Prize 2024 Technical Report~\cite{arcprize2024technicalreport}.

Shortly following the conclusion of the competition in late 2024, a preview version of OpenAI's ``o3'' model, in private testing, demonstrated remarkable performance on the ARC-AGI-1 Semi-Private Evaluation dataset. Using significant computational resources at test time, this model achieved scores of 76\% (low-compute; estimated cost: \$200 per task) and 88\% (high-compute; estimated cost: \$20,000 per task)~\cite{o3arcagi}, surpassing the nominal human baseline for the first time. Subsequent publicly released versions of o3 showed lower, albeit still strong, performance (e.g., 53\% for o3-medium on the Semi-Private set)~\cite{o3analysis}, the preview result indicated that ARC-AGI-1, while still challenging, was approaching saturation under the test-time adaptation paradigm using very large amounts of compute power.

ARC Prize 2024 provided several key insights. Firstly, it showed that progress on ARC-AGI requires moving beyond simply scaling the formerly prevailing deep learning paradigm. Notably, test-time adaptation methods emerged as a necessary strategy used by all top teams. The competition successfully incentivized the exploration and open-sourcing of these new directions, confirming ARC-AGI's value in pushing AI research towards more general and flexible reasoning capabilities. It also demonstrated the need for an upgraded and more challenging version of ARC-AGI to continue driving progress towards artificial general intelligence.

\section{Limitations of ARC-AGI-1}

Previous competitions have served to highlight several key limitations~\cite{arcprize2024technicalreport} of ARC-AGI-1:

\textbf{Task susceptibility to non-generalizable strategies}

In the inaugural 2020 competition, the top-performing individual submission achieved a score of 20\%. However, a subsequent meta-analysis, aggregating the unique tasks solved across all submissions from that year, found that 49\% of the Private Evaluation set was successfully solved by at least one team. Crucially, the dominant techniques employed by these successful submissions were reported to be variations of brute-force program search.

This suggests that nearly half of the ARC-AGI-1 Private Evaluation tasks might be vulnerable to computationally intensive, exhaustive search methods without requiring the kind of efficient abstract reasoning abilities we consider central to AGI. Although the remaining tasks ($\sim$50\%) have proven sufficiently challenging to resist current methods (leaving the benchmark unsolved), the significant fraction susceptible to brute force considerably dilutes the benchmark signal. It may reward computational power over the development of more general cognitive architectures. A robust AGI benchmark should ideally minimize susceptibility to such non-generalizable solution strategies.

\textbf{Lack of reliable first-party human testing data}

While the original ARC-AGI private tasks were confirmed solvable by humans (initial testing by two individuals yielded scores of 97\% and 98\%), and while some third-party studies have investigated the performance of Mechanical Turkers on the public datasets (such as NYU's "Fast and flexible: Human program induction in abstract reasoning tasks"~\cite{nyu2021} and "H-ARC: A Robust Estimate of Human Performance on the Abstraction and Reasoning Corpus Benchmark"~\cite{nyu2024}), there was no official first-party human baseline score derived under consistent conditions for the hidden tasks used in competitions.

The lack of such a human baseline makes a definitive understanding of human performance challenging. Variations in participant pools, motivation, time constraints, and interfaces used in third-party studies can introduce variability compared to a potential standardized first-party protocol.

\textbf{Saturation below human-level fluid intelligence}

Empirically, humans at the higher end of the intelligence distribution could solve over 97\% of ARC-AGI-1 tasks without much effort, which means that the benchmark saturated well before it could capture the full spectrum of human fluid intelligence.

\textbf{Inconsistent difficulty distribution}

Based on empirical scores, ARC-AGI-1 exhibits potential inconsistencies in difficulty distribution across its different data subsets (i.e., the Public Evaluation set has generally been found to be easier than the the Private Evaluation set). If the subsets do not represent comparable draws from an underlying task difficulty distribution, it becomes challenging to reliably interpret scores.

\textbf{Risk of information leakage}

The same 100 Private Evaluation tasks have been reused and unchanged across all four major ARC-AGI-1 competitions (2020-2024) to provide intermediate leaderboard feedback to participants. Consequently, an estimated 10,000 scores derived from performance on this hidden set have been disclosed over time. While individual scores offer limited insight, each data point represents a potential channel for information leakage, however small. Over thousands of iterations across numerous teams, participants can implicitly or explicitly infer characteristics of the hidden tasks based on how score changes correlate with submission modifications. This cumulative feedback loop creates a substantial risk that models become progressively tuned to the specific idiosyncrasies of these 100 tasks, rather than developing truly generalizable reasoning capabilities. Performance improvements on the leaderboard might therefore reflect adaptation to the specific test set rather than genuine advances in abstract problem-solving.

\section{Goals of ARC-AGI-2}

In response to these limitations, we started developing ARC-AGI-2 in late 2021. This updated version of the benchmark aims to be a continuation of ARC-AGI-1, while fully addressing the issues listed above.

Our key goals in this update were as follows:

\begin{enumerate}

    \item \textbf{Same fundamental principles.} Maintain the fundamental principles of ARC-AGI-1: each task is unique and cannot be memorized in advance, all tasks require only elementary Core Knowledge, and all tasks seek to adhere to the ``easy for humans, hard for AI'' design guideline.

    \item \textbf{Same format.} Retain the established and well-understood task format of ARC-AGI-1, with tasks defined via input-output pairs of grids (sized from 1x1 to 30x30) composed of cells with 10 possible discrete values (colors). This ensures familiarity for researchers and facilitates the reuse of existing tooling and visualization methods.

    \item \textbf{Less brute-forcible.} Intentionally design tasks to minimize susceptibility to naive or computationally intensive brute-force program search techniques, since such tasks provide no signal with regard to AGI progress. This shifts focus further towards efficient adaptation.
    
    \item \textbf{Extensive first-party human testing.} Run large-scale live testing with a diverse population of human participants. This provides robust first-party data on human solvability rates, perceived difficulty, and solution strategies, establishing reliable human baseline performance metrics (including accuracy and, potentially, efficiency measures like time or cost) against which AI systems can be rigorously compared.

    \item \textbf{Wider useful ``signal bandwidth''.} Provide a wider useful range of scores to measure AI capabilities. By including tasks that span a carefully considered spectrum of difficulty, while remaining generally accessible to humans, and by reducing the number of tasks solvable by near-trivial means, ARC-AGI-2 should better differentiate between systems possessing varying levels of fluid reasoning ability.

    \item \textbf{Calibrated human-facing difficulty across subsets.} Curate each subset (Public Evaluation, Private Evaluation, Semi-Private Evaluation) such that they are drawn from demonstrably similar distributions in terms of human solvability and perceived difficulty, ensuring that performance on one set is reliably predictive of performance on others.

\end{enumerate}

\section{Human-facing calibration testing}

\subsection{Protocol}

We conducted testing on potential ARC-AGI-2 tasks with human volunteers in a controlled environment. Participants worked individually on computers in a conference room setting, with a maximum of 34 participants at any given time. Assigned tasks were randomized and presented through a custom user interface with minimal controls. Participants completed a short survey and interface tutorial prior to being assigned tasks. Participants received a base compensation of \$115-150 for participation in a 90-minute test session, plus a \$5 incentive reward per correctly completed task. Three testing sessions were held between November 2024 and May 2025.

\subsection{Demographics}

\begin{figure}[h]
\centering
\includegraphics[width=0.8\linewidth]{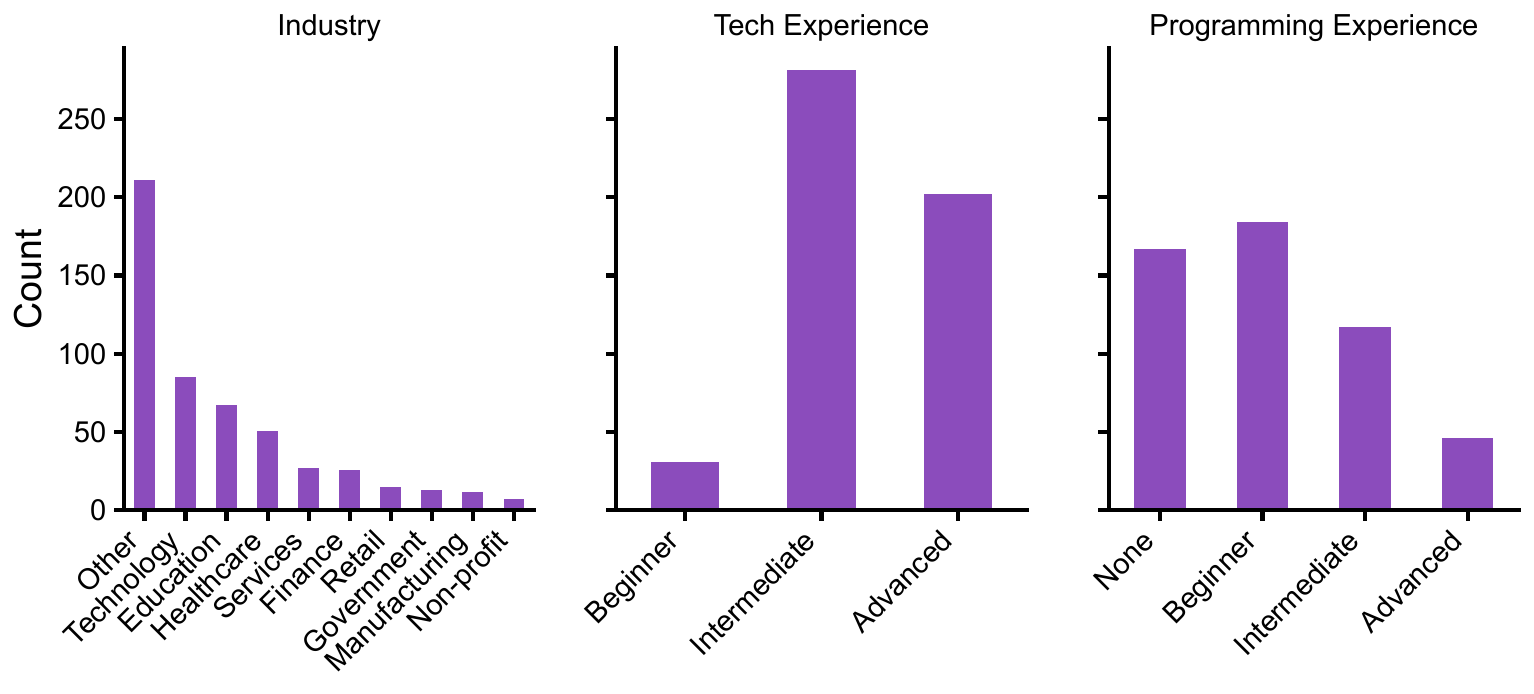}
    \caption{Study participants by self-reported industry and experience}
\label{fig:demographics}
\end{figure}

Study participants came from diverse professional backgrounds, with a wide variation in self-reported experience in technology, programming, math, and puzzle solving (partially shown in Figure \ref{fig:demographics}).

\subsection{Results}
\label{sec:human_testing_results}

Most (68\%) of the tasks tested contained a single test pair, while the remainder had two (29\%), three (3\%), or four (<1\%) test pairs. Solutions were considered fully correct only when all test pairs per task were solved and partially correct when at least one test pair was solved.

We define an "attempt" as any task view lasting longer than 5 seconds.

\begin{figure}[htbp]
\centering
\includegraphics[width= 0.6\linewidth]{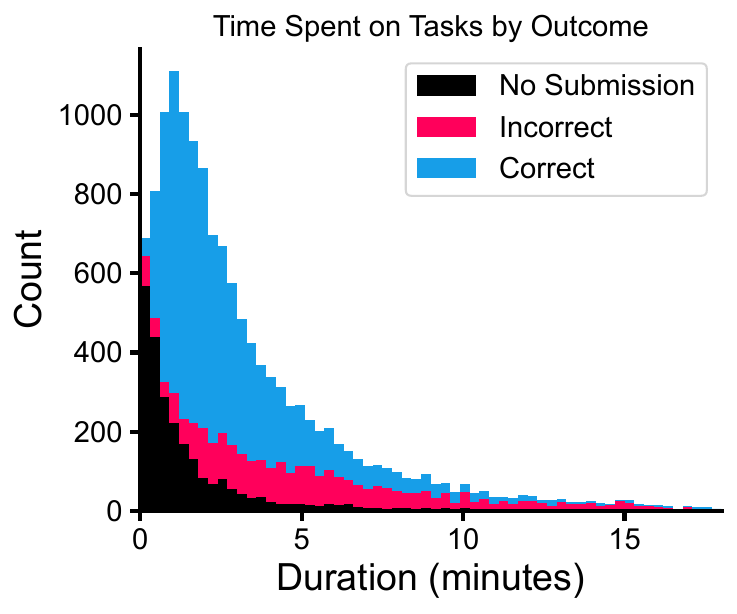}
    \caption{Outcome and time spent on all task test pair attempts across all participants}
    \label{fig:task_outcomes}
\end{figure}

Using these definitions, we recorded 407 unique participants in 515 sessions attempting 1,848 unique task test pairs (many of which were not included in the final ARC-AGI-2). This resulted in 13,405 total test pair attempts, of which 8,277 (62\%) were successfully solved. The median time spent on attempted test pairs was 2.3 minutes, while successfully completed tasks required a median of 2.2 minutes (Figure \ref{fig:task_outcomes}).

\subsection{Performance}

\begin{figure}[h]
\centering
\includegraphics[width= 0.6\linewidth]{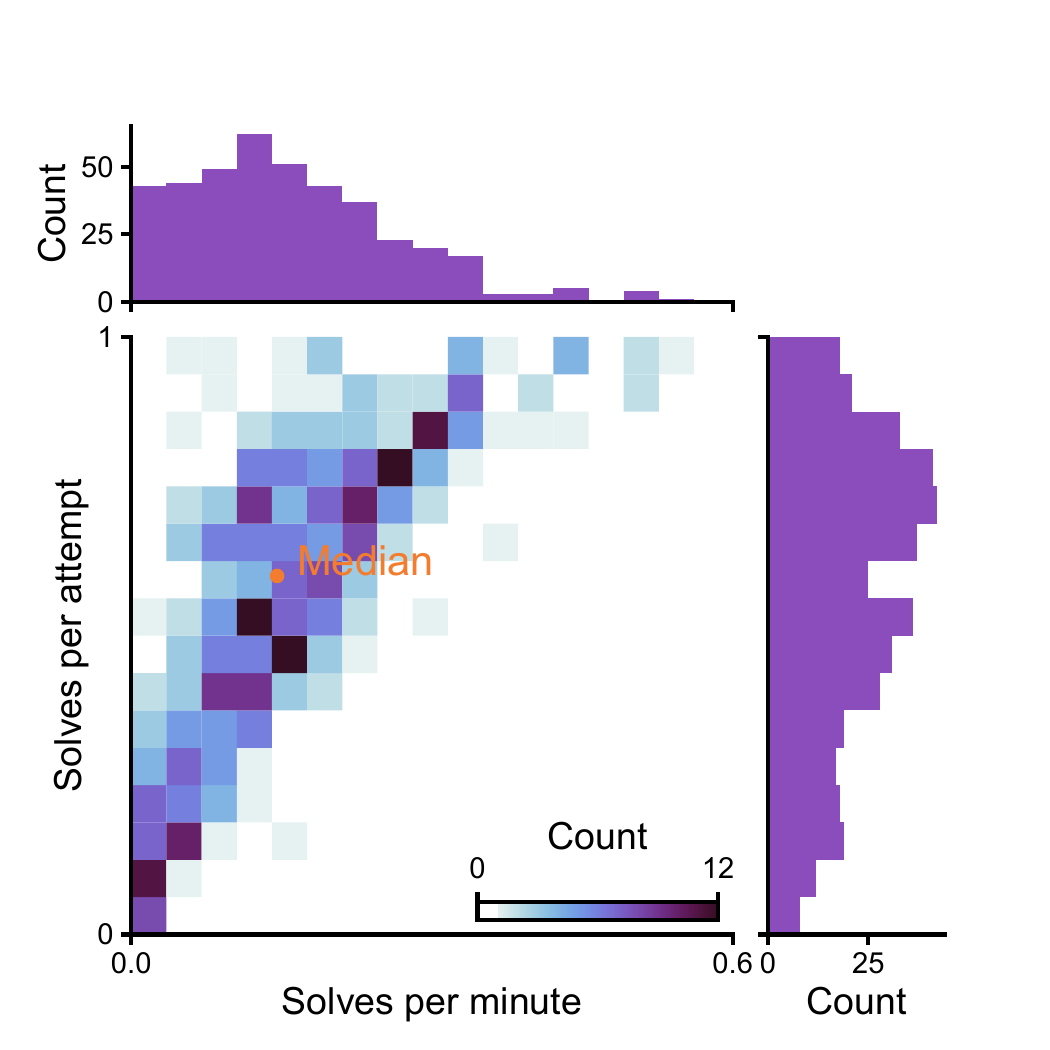}
    \caption{Task solving speed and accuracy of participants}
\label{fig:speed_and_accuracy}
\end{figure}

Participant performance varied considerably in both speed (solved tasks per minute) and accuracy (solved tasks per attempt), and, in general, the two showed a positive correlation (Figure \ref{fig:speed_and_accuracy}). 

Examining the activity of individual participants during a session, we observed that most participants proceeded sequentially through tasks as instructed, with occasional backtracking to reexamine a task they'd previously not solved.

None of the self-reported demographic factors recorded for all participants---including occupation, industry, technical experience, programming proficiency, mathematical background, puzzle-solving aptitude, and various other measured attributes---demonstrated clear, statistically significant relationships with performance outcomes. This finding suggests that ARC-AGI-2 tasks assess general problem-solving capabilities rather than domain-specific knowledge or specialized skills acquired through particular professional or educational experiences.

\section{Task selection process}

\paragraph{Initial screening and task inclusion.}
Candidate tasks originated from two sources: newly authored tasks produced specifically for ARC-AGI-2 by ARC Prize Foundation staff and partners, along with previously unused reserves from previous ARC-AGI iterations. Because increasing the number of tasks assigned during human testing is comparatively inexpensive, we deliberately over-generated tasks, anticipating attrition during curation. Tasks from the ARC-AGI-1 Public Training set were excluded from testing.

A task advanced from human testing only if at least two independent participants solved \emph{one or more} sub-pairs within their first two attempts.

\paragraph{Difficulty calibration of task subsets.}
For each task, we computed the proportion of participants achieving full correctness and treated this as an empirical difficulty index. Tasks were grouped into public, semi-private, and private sets such that the mean human accuracy differed by $\leq$ 1 percentage point across partitions. Newly authored tasks were preferentially allocated to private sets, whereas tasks previously publicly available remained public.

\paragraph{Redundancy detection.}
A custom review interface presented all qualified tasks for visual comparison. Two tasks were deemed redundant when it was judged that a single programmatic solution would likely generalize to solve both. Flagged pairs underwent consensus review, which resulted in non-overlapping final tasks.

\paragraph{Training subset.}
Tasks that were easily solved by most test takers were placed in the Public Training set. This set was not difficulty-calibrated. The Public Training set is not an evaluation set, rather, it serves as a general repository for validated tasks intended for model training and for demonstrating the ARC-AGI format. As a result, the Public Training set contains a range of difficulty levels. Additionally, not all tasks in this set were fully human-tested.

\begin{figure}[h]
\centering
\includegraphics[width= 0.55\linewidth]{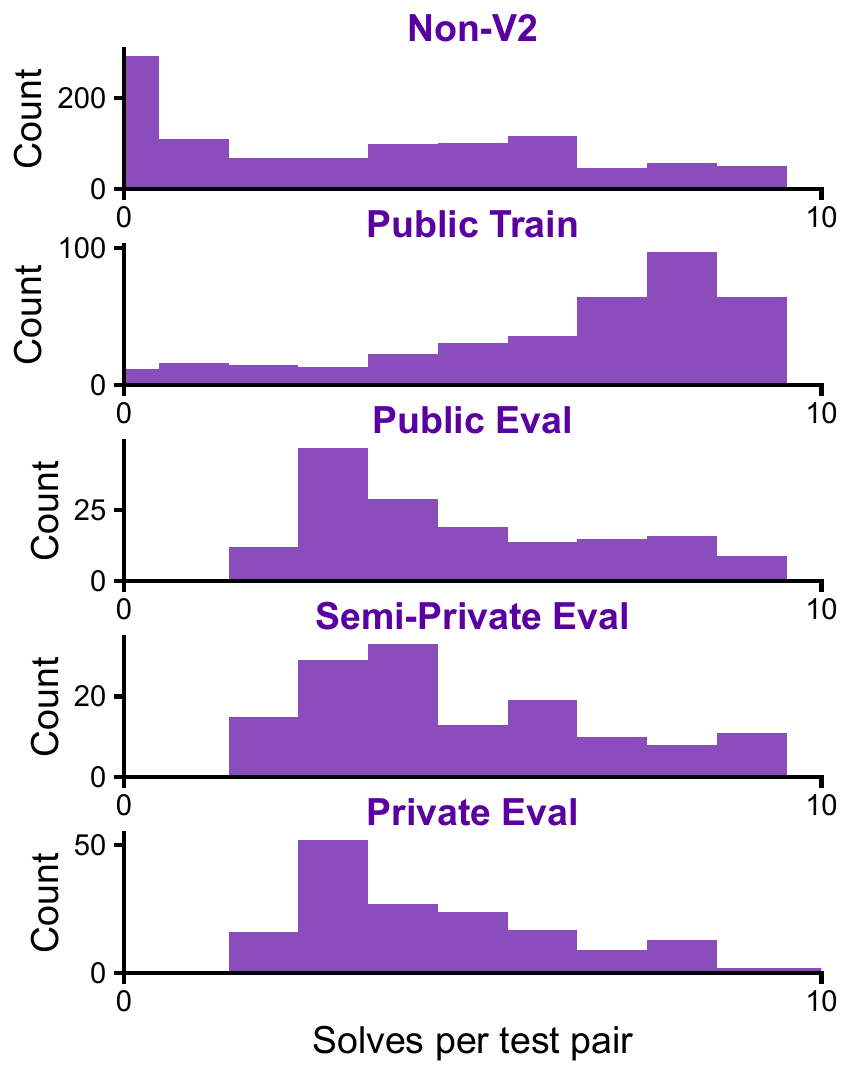}
    \caption{Number of human participants that correctly solved ARC-AGI-2 task test pair subsets. Note that Public Eval, Semi-Private Eval and Private Eval filtered out tasks which less than 2 people solved.}
\label{fig:v2_task_difficulty}
\end{figure}

\paragraph{Final validation.}
All tasks underwent a two-layer validation process. Externally, two independent human testers completed every task in the ARC-AGI-2 Public Evaluation, Semi-Private, and Private sets~(Figure \ref{fig:v2_task_difficulty}). This provided a first-pass confirmation of solvability. Internally, additional tasks were examined and required to be solved by an additional reviewer beyond the original author.

These protocols prioritized test pair correctness. Minor cell-level inconsistencies were discovered in a handful of training pairs. While these errors were not intentional and showed no impact on test-pair solvability (humans knew to disregard these noisy cells), they were corrected wherever found to preserve task aesthetics and logical consistency.

After selection and validation, 100\% of ARC-AGI-2 tasks were solved by at least two independent, non-expert human testers drawn from the general public, with each task attempted by between 2 and 10 humans. Aggregating results by task, 75\% of human attempts were successfully completed. Aggregated across all human evaluations and attempts, 66\% of attempted test pairs were successfully completed. Together, these results establish that all tasks in ARC-AGI-2 are human-solvable.

\section{State-of-the-art}

Baseline model performance on ARC-AGI-2 was generated with the publicly available \texttt{Model Baseline} repository \cite{arcprize:model_baseline}. Models were evaluated on the \emph{Semi-Private} Evaluation set.

Complete updated scores are hosted on the official ARC-AGI Leaderboard \cite{arcprize:leaderboard}.

\begin{table}[h]
    \centering
    \label{tab:sota}
    \begin{tabular}{lcc}
        \toprule
        \textbf{Model} & \textbf{ARC-AGI-1} & \textbf{ARC-AGI-2} \\
        \midrule
        o3-mini (High)              & 34.5\% & 3.0\%   \\
        o3 (Medium)                 & 53.0\% & 3.0\%   \\
        ARChitects (ARC Prize 2024) & 56.0\% & 2.5\%   \\
        o4-mini (Medium)            & 41.8\% & 2.4\%   \\
        Icecuber (ARC Prize 2020)   & 17.0\% & 1.6\%   \\
        o1-pro (Low)                & 23.3\% & 0.9\%   \\
        Claude 3.7 (8K)             & 21.2\% & 0.9\%   \\
        \bottomrule
    \end{tabular}
    \caption{Top performing models on the Semi-Private Evaluation set as of May 14, 2025}
\end{table}

While scores above 0\% indicate that a model has solved at least one task, ARC-AGI-2 accuracies below 5\% are generally not treated as meaningful, as they likely result from noise-level heuristics or incidental pattern fits. In our experience, consistent signal begins to emerge only once performance exceeds the 5\% threshold.

\section{What makes ARC-AGI-2 more challenging?}

\subsection{Key design changes}

Many ARC-AGI-1 tasks could often be solved almost instantaneously by human test-takers without requiring significant cognitive effort. In contrast, all tasks in ARC-AGI-2 require some amount of deliberate thinking --- for instance, the average time for task completion among human test takers in our sample was 2.7 minutes. This increased difficulty stems from several key design changes.

First, an elementary change is that ARC-AGI-2 tasks are \textit{more unique} --- while a number of ARC-AGI-1 tasks had some amount of overlap with patterns that could be found elsewhere (due to their simplicity), every ARC-AGI-2 task is, to the best of our knowledge, entirely novel.

Second, ARC-AGI-2 tasks are \textit{more complex} in terms of their information content --- they generally feature larger grids, more objects per grid, and more concepts per task. Any attempt at compressing ARC-AGI-2 tasks would result in more bits per task than for ARC-AGI-1.

Lastly, a major focus with ARC-AGI-2 is to investigate deeper levels of \textit{compositional generalization} --- the ability to combine known rules or concepts in novel ways. This often takes the form of multi-rule compositional reasoning, multi-step compositional reasoning, contextual rule application, and in-context symbol definition. Below, we review these design rules, show qualitative examples, and discuss what makes them challenging for AI systems.

\subsection{Designing for compositional generalization: examples}

\begin{figure}[h]
  \centering
  \includegraphics[width=0.6\textwidth]{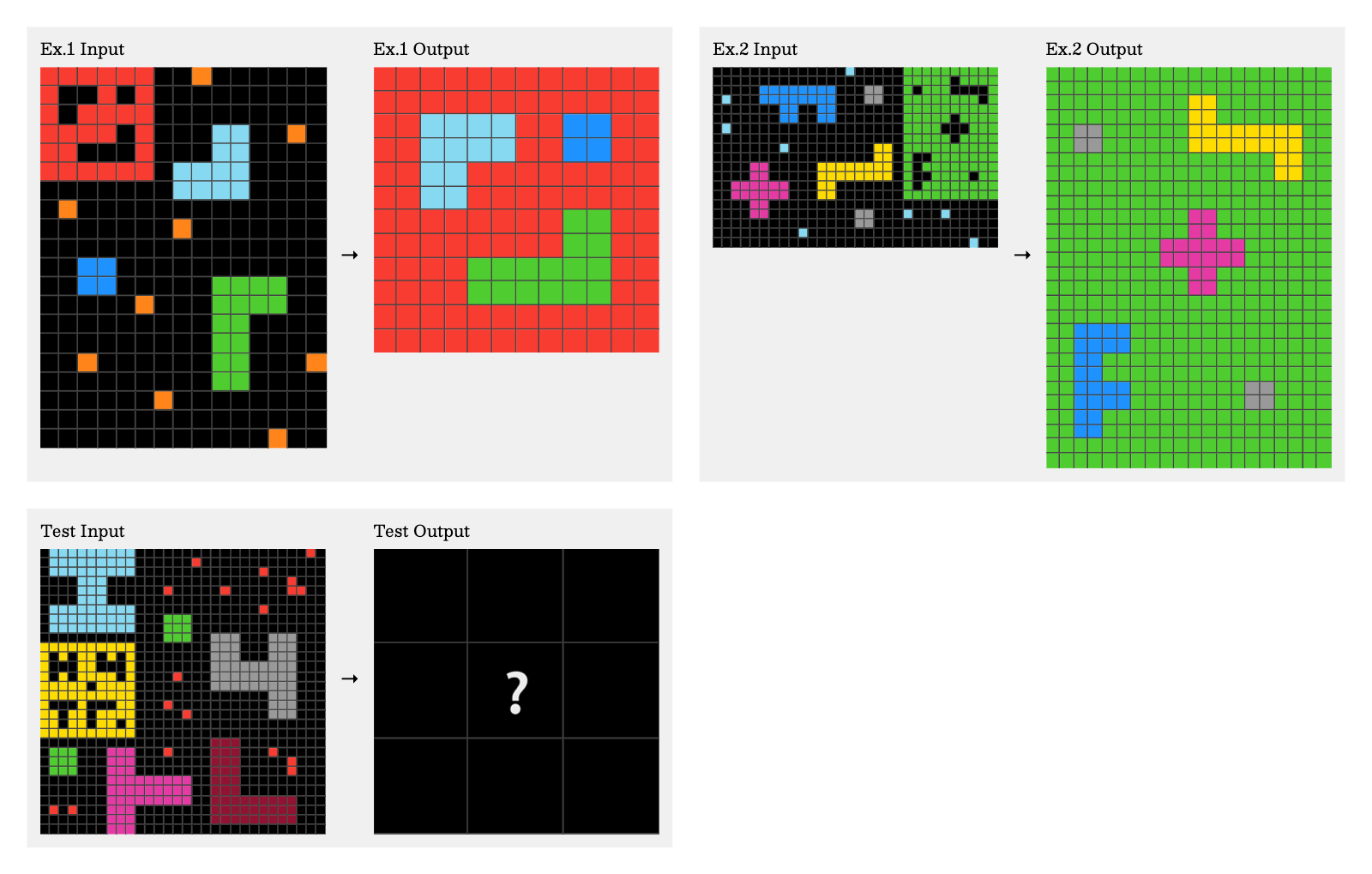}
  \caption{ARC-AGI-2 task id: 898e7135}
  \label{fig:task-898e7135}
\end{figure}

\textbf{Multi-rule compositional reasoning.} While most ARC-AGI-1 tasks could be solved by identifying and applying a single high-level transformation (e.g., ``objects fall down''), ARC-AGI-2 tasks seek to feature multiple simultaneous rules, often interacting with each other. For instance, in Figure \ref{fig:task-898e7135}, one must crop the input grid to the rectangular framed area, rescale the colored objects, and place the rescaled objects into corresponding holes of the same shape within the framed area.

\begin{figure}[h]
  \centering
  \includegraphics[width=0.6\textwidth]{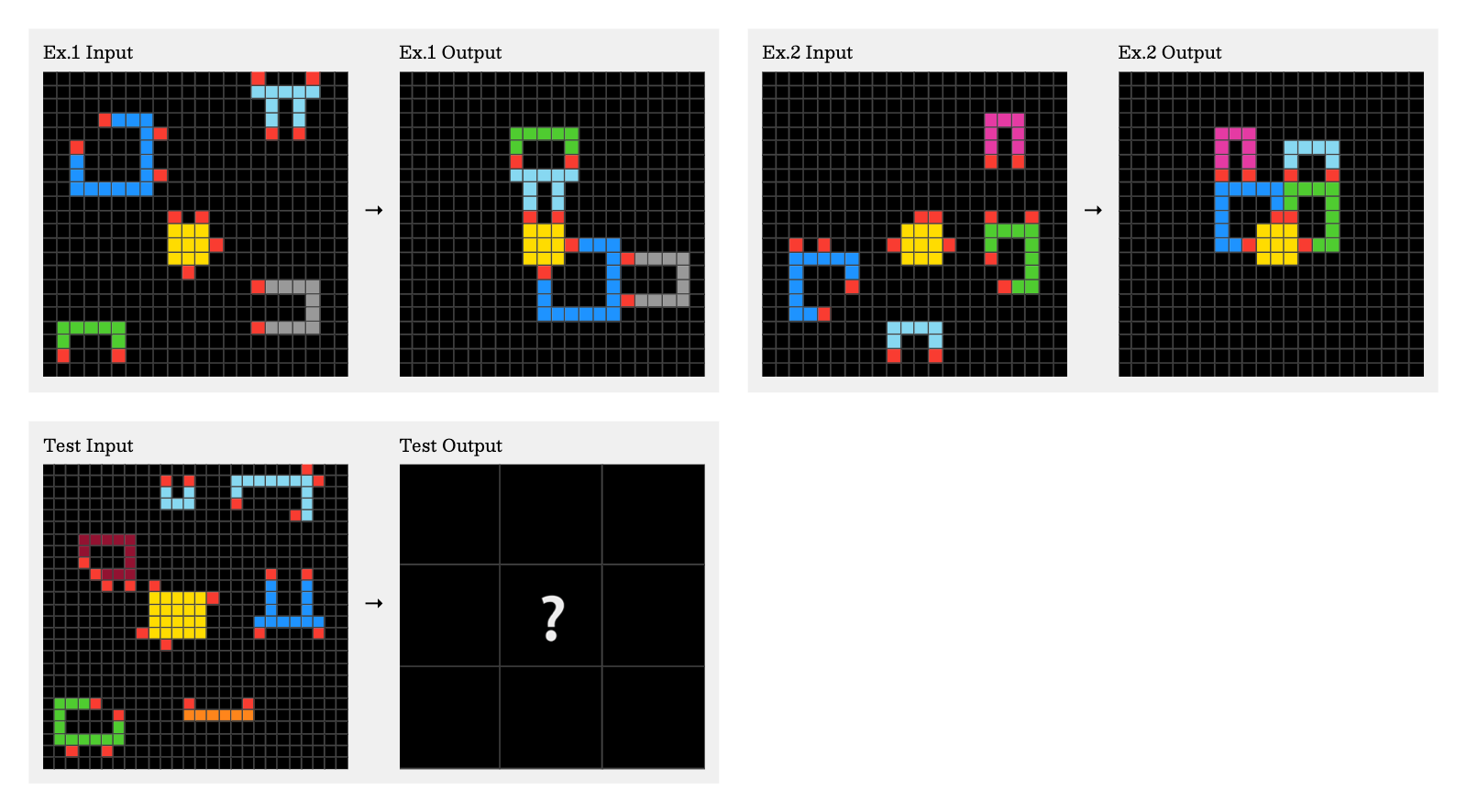}
  \caption{ARC-AGI-2 task id: cbebaa4b}
  \label{fig:task-cbebaa4b}
\end{figure}

\textbf{Multi-step compositional reasoning.} Many ARC-AGI-2 tasks require the sequential application of a rule where the state after step N depends directly on the outcome of step N-1. An example might involve placing objects iteratively, where the correct position and orientation of the next objects are determined by the placement of the previous ones (as in Figure \ref{fig:task-cbebaa4b}). It is virtually impossible to predict the position of object $N + 1$ without executing the previous $N$ steps.

\begin{figure}[h]
  \centering
  \includegraphics[width=0.6\textwidth]{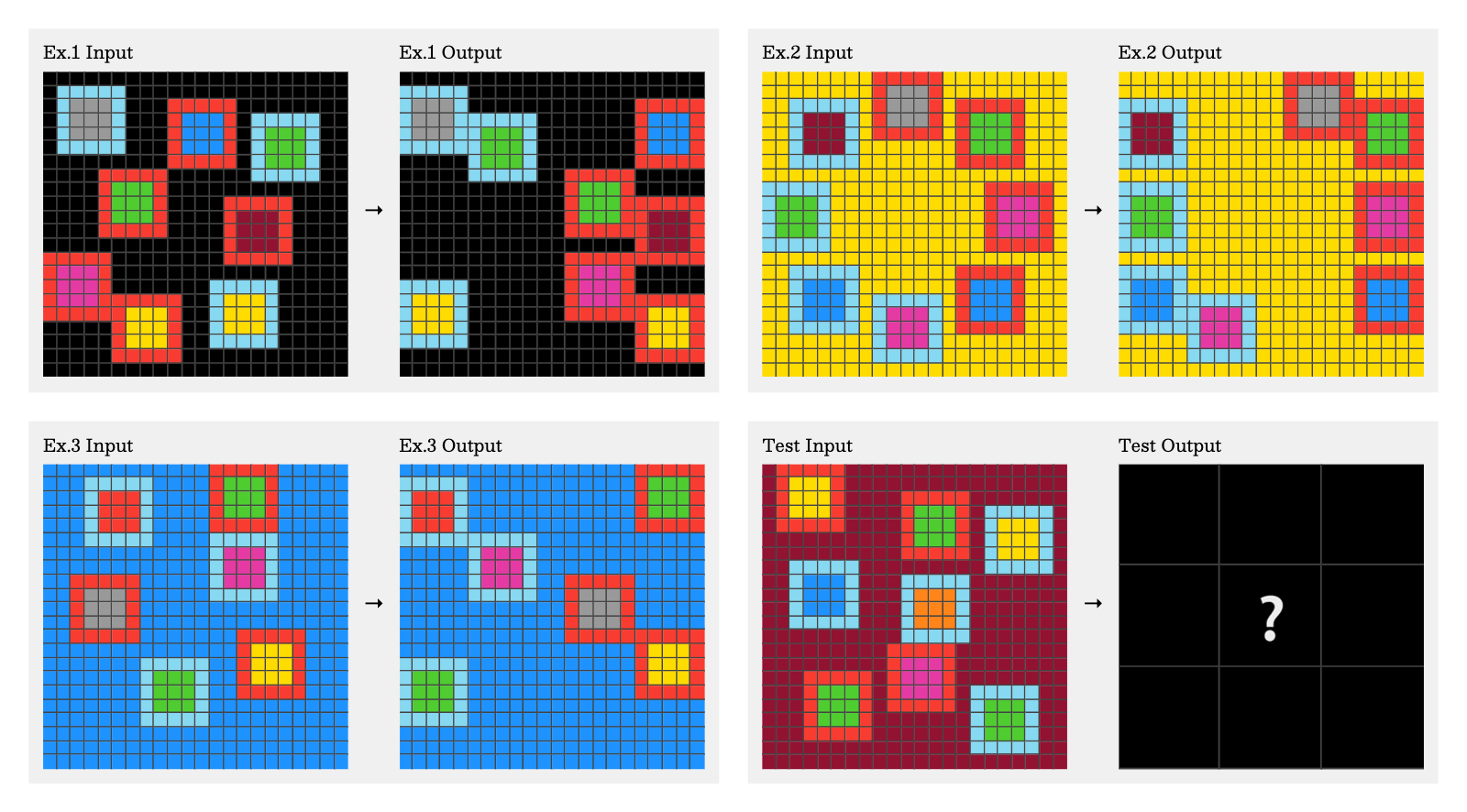}
  \caption{ARC-AGI-2 task id: b5ca7ac4}
  \label{fig:task-b5ca7ac4}
\end{figure}

\textbf{Contextual rule application.} ARC-AGI-2 features tasks where test-takers must not only identify the core transformation rule but also understand how its application is modulated by specific contextual elements within the grid. This adds an extra hop to the reasoning chain, often involving a form of control flow. For example, a task might involve isolating shapes and stacking them to the side as in Figure \ref{fig:task-b5ca7ac4}, but the decision of which side (e.g., left or right) depends on a contextual cue, such as the color of the shape's outline. While current systems might identify the basic shape-stacking operation, correctly interpreting and applying the contextual gating mechanism remains a significant challenge. This requires composing the transformation rule with a selection or conditional rule derived from the context.

\textbf{In-context symbol definition.} Many ARC-AGI-2 tasks feature ``symbols'' (objects that stand for something other than themselves) whose meaning is defined within the task, such as in Figure \ref{fig:task-e3721c99}, where colored rectangles with holes encode the color to use for shapes that have the same number of holes. This kind of on-the-fly symbolic assignment has been, in our experience, a major challenge for frontier AI systems.

\section{ARC Prize 2025}
    \subsection{Kaggle competition}
    The global ARC Prize competition returned in 2025 to offer \$1,000,000 USD in prizes to accelerate research on open source progress towards completing ARC-AGI. Its goal is to inspire AI researchers to work on new ideas and approaches through the specific challenge of open-sourcing a solution to ARC-AGI. The competition is aimed at individual researchers and small teams---driven by curiosity, inspired by complexity, and committed to the rigorous pursuit of genuine general intelligence.

    The \$1,000,000 prizes are awarded as follows:
    
    \begin{enumerate}
        \item \textbf{Grand Prize} (\$700,000): Awarded to the first team(s) to get $\geq$85\% accuracy on the hidden ARC-AGI-2 Private Evaluation set.
        \item \textbf{Annual Progress Prizes} (\$125,000): Split between the Top Score (\$50K) and Paper Prize (\$75K) and guaranteed to be awarded in 2025.
        \item \textbf{To-Be-Determined Prizes} (\$175,000): An additional prize pool reserved for rewarding outstanding achievements.
    \end{enumerate}

    \textbf{Evaluation protocol.} Each submission is executed offline inside Kaggle’s secure sandbox server environment on four NVIDIA L4 GPUs. Within a single 12-hour wall-clock window, submission code must solve 240 previously unseen ARC-AGI-2 tasks, 120 in the Semi-Private Evaluation set and 120 in the Private Evaluation Set, without any internet access (to prevent data leakage).
    
    Semi-Private Evaluation accuracy is reflected on the public leaderboard after each submission; Private Evaluation accuracy remains hidden until entrants open-source their solutions after the competition. The final standings are determined by the Private Evaluation set score, which is calculated once the competition has closed.

    \textbf{Timeline.} ARC Prize 2025 launched on March 24, 2025. Final submissions are due by November 3, 2025. Papers are due by November 9, 2025.

    \subsection{Public Leaderboard}
    The public ARC-AGI Leaderboard \cite{arcprize:leaderboard} offers a snapshot of how frontier closed-weight models perform on the benchmark. While these submissions do not qualify for ARC Prize 2025 awards, they demonstrate what publicly accessible systems can currently achieve.

    The leaderboard is organized as a 2×2 matrix with axes for cost per task and score. Mapping models into this space reveals trade-offs between efficiency (with cost as a proxy) and task performance.
    
    ARC Prize also restricts which companies and models are eligible for testing. The public testing policy is provided on the ARC Prize website at arcprize.org/policy\cite{arcprize:policy}.

\section{Conclusions}

ARC-AGI-2 is a significant evolution of the original benchmark, addressing known limitations while preserving its core principles and format. It introduces tasks with greater complexity and uniqueness, specifically designed to resist brute-force methods and better evaluate compositional generalization. Extensive human calibration ensures that tasks remain feasible for humans and are accurately difficulty-calibrated.

\section{Acknowledgments}

We are grateful to Coral Lovci and Breakthrough Properties for generously providing The Clubhouse at Torrey Heights in San Diego, which hosted the majority of our testing sessions. We also sincerely thank Lizbeth Dargi, Andre Vu, and Roxana Behjat for their support in administering and facilitating human testing.

We also extend our gratitude to select members of the Ndea team for contributing candidate tasks considered for human testing and subsequent inclusion in ARC-AGI-2.

\bibliographystyle{plain}
\bibliography{references_new}

\end{document}